\documentclass[letterpaper, 10 pt, conference]{ieeeconf}  

\IEEEoverridecommandlockouts                              

\usepackage{multicol}
\usepackage[bookmarks=true]{hyperref}
\usepackage{amsmath}
\usepackage{mathtools}
\usepackage{algorithm}
\usepackage{algpseudocode} 
\usepackage{datetime}
\usepackage{enumerate}
\usepackage{tikz}
\usepackage{xcolor}
\usepackage{amsfonts}
\usepackage[font=small]{subcaption}
\usepackage[font=small]{caption}
\usepackage{relsize}
\usepackage{tabularx}
\usepackage{lipsum}
\usepackage{array}
\usepackage{LaTeXPackages/abkuerzungen}
\usepackage{comment}

\pgfdeclarelayer{background}
\pgfsetlayers{background,main}

\definecolor{lavander}{cmyk}{0,0.48,0,0}
\definecolor{violet}{cmyk}{0.79,0.88,0,0}
\definecolor{burntorange}{cmyk}{0,0.52,1,0}
\definecolor{hlblue}{cmyk}{0.5.0,0.52,0.0,0}
\definecolor{grey}{cmyk}{0.5.0,0.5,0.5,0}

\def\lav{lavander!90}
\def\oran{orange!30}

\tikzstyle{peers}=[draw,circle,grey,bottom color=grey,
                  top color= white, text=black,minimum width=10pt]
\tikzstyle{superpeers}=[draw,circle,burntorange, left color=\oran,
                       text=black,minimum width=20pt]
\tikzstyle{candidate}=[draw,circle,hlblue, bottom color=hlblue, top color=white,
                       text=violet,minimum width=20pt]
\tikzstyle{legendsp}=[rectangle, draw, burntorange, rounded corners,
                     thin,bottom color=\oran, top color=white,
                     text=burntorange, minimum width=2.5cm]
\tikzstyle{legendp}=[rectangle, draw, violet, rounded corners, thin,
                     bottom color=\lav, top color= white,
                     text= violet, minimum width= 2.5cm]
\tikzstyle{legend_general}=[rectangle, rounded corners, thin,
                           burntorange, fill= white, draw, text=violet,
                           minimum width=2.5cm, minimum height=0.8cm]
\tikzstyle{edge} = [draw,thick,-]
\tikzstyle{weight} = [font=\small]

\title{\LARGE \bf
Appearance-Based Landmark Selection for Efficient Long-Term Visual Localization
}

\author{Mathias B\"{u}rki, Igor Gilitschenski, Elena Stumm, Roland Siegwart and Juan Nieto
\\ Autonomous Systems Lab, ETH Z\"{u}rich
\\ {\tt\footnotesize \{firstname.lastname\}@mavt.ethz.ch}
}

\begin{document}
\bstctlcite{IEEEexample:BSTcontrol}

\newcommand{\poseguesslong}[1]{\mathcal{T}_{{WB}_#1}}
\newcommand{\poseguess}[1]{\mathcal{T}}
\newcommand{\pastobslong}{\mathcal{V}_{k-1}}
\newcommand{\pastobs}{\mathcal{V}}
\newcommand{\candidate}{l}

\newcommand{\likelihood}{p(\pastobs \mid l_i)}
\newcommand{\voteprob}{p(l_j \mid l_i)}
\newcommand{\lochist}{\mathbf{Z}}

\maketitle
\thispagestyle{empty}
\pagestyle{empty}

\begin{abstract}

In this paper, we present an online landmark selection method for distributed long-term visual localization systems in bandwidth-constrained environments.
Sharing a common map for online localization provides a fleet of autonomous vehicles with the possibility to maintain and access a consistent map source, and therefore reduce redundancy while increasing efficiency.
However, connectivity over a mobile network imposes strict bandwidth constraints and thus the need to minimize the amount of exchanged data.
The wide range of varying appearance conditions encountered during long-term visual localization offers the potential to reduce data usage by extracting only those visual cues which are relevant at the given time.
%
Motivated by this, we propose an unsupervised method of adaptively selecting landmarks according to how likely these landmarks are to be observable under the prevailing appearance condition. 
The ranking function this selection is based upon exploits landmark co-observability statistics collected in past traversals through the mapped area. 
Evaluation is performed over different outdoor environments, large time-scales and varying appearance conditions, including the extreme transition from day-time to night-time, demonstrating that with our appearance-dependent selection method, we can significantly reduce the amount of landmarks used for localization while maintaining or even improving the localization performance.
\end{abstract}

\section{Introduction}
\label{sec:introduction}

A fundamental problem to be tackled to enable fully autonomous driving is the cooperation and coordination among multiple vehicles, including sharing and exchanging information.
This will be a key aspect for the success in coping with the complexity, variability, and volatility of typical urban environments.
Especially for the task of localization and mapping, sharing and maintaining a common map offers a high potential for reducing data redundancy and for providing timely up-to-date maps. 
Vehicles will be required to exchange data among themselves, and/or with a common cloud-based map-service. 
Since bandwidth on mobile data networks is a scarce resource, it is pivotal to minimize the amount of information exchanged. 
This is particularly important for visual localization and mapping, where appearance variations generate the need to store many different representations for each location \cite{churchill2013experience}, \cite{muhlfellner2015summary}.

\begin{figure}
\includegraphics[width=0.5\textwidth]{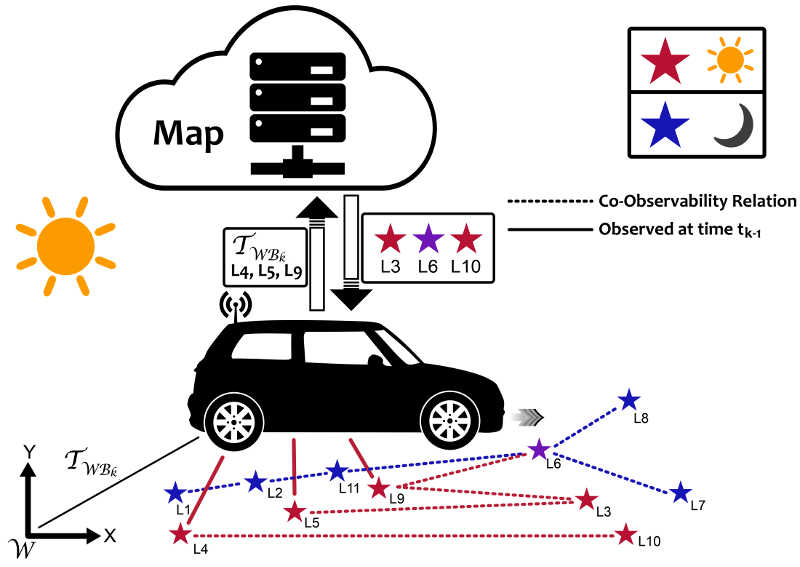}
\caption{A schematic illustration of a distributed visual localization system using online landmark selection. A vehicle continually receives selective visual landmarks for localization from a cloud-based map server during operation, while transmitting back its pose and information about recently observed landmarks. In the depicted situation, landmarks $L4$ $L5$ and $L9$ have recently been observed, therefore their IDs, plus a rough initial estimate of the vehicle's current pose are transmitted to the map server. In response, a subset of relevant landmarks, consisting of $L3$, $L6$ and $L10$ are transmitted back to the vehicle and used for subsequent localization.
\label{fig:schematic}}
\vspace{-3mm}
\end{figure}	


To approach this problem, we propose an online landmark selection method which - without losing localization performance - is able to significantly reduce the amount of data exchanged between the vehicle and its map source. 


The general principle of the method can be summarized as follows: 
\begin{itemize}
\item The prevailing appearance condition of the environment is inferred from landmarks observed in recent localization attempts during a traversal through the mapped area.
\item Using this information, all landmarks in a spatially local neighborhood (the candidate landmarks) are ranked according to how likely they are to be observed in subsequent localization attempts along the same traversal.
\item A selected subset of top-ranked candidate landmarks is transferred back to the vehicle and used for localization.
\end{itemize}

Localization can then be performed on the vehicle, based on the selection of suitable landmarks, which is computed on a remote map server and sent to the vehicle. The data exchanged during each localization attempt consists of a rough initial estimate of the vehicle's current pose, references (e.g. IDs) to recently observed landmarks, and a reduced set of selected landmarks. A schematic illustration of this distributed localization paradigm can be found in figure \ref{fig:schematic}.

The key for an effective landmark selection is the ranking process. In our approach, this ranking is performed in an unsupervised manner, based on co-observability statistics between a candidate landmark and a set of recently observed landmarks collected in past traversals through the same area. 
\newline

The main contribution of the proposed approach is the derivation of an online landmark selection method based on co-observability statistics. The motivation for this work is the need for a localization and mapping strategy, that can deal with the bandwidth-constrained settings found in distributed systems operating in changing environments. In particular, our approach provides the following features:
\begin{itemize}
\item Efficient and accurate localization using only an appearance-dependent subset of landmarks inferred at runtime in an unsupervised manner.
\item The size of the selected subset is adaptable to prevailing bandwidth restrictions. 
\item Computational demands on the vehicle are reduced by significantly cutting down the amount of input data used for localization.
\end{itemize}

We evaluate our approach in two complementary scenarios. 
In the first scenario, our landmark selection method is evaluated in a long-term experiment on an outdoor parking-lot, covering day-time conditions observed over the time frame of one year.
In the second scenario, our method is evaluated in a small city environment, covering extreme appearance changes from day-time to night-time over the time frame of one day. 
The results validate our approach by showing that with our landmark selection method, we significantly reduce the amount of data exchanged between the vehicle and the map, while maintaining comparable or even better localization performance than if all data is used. 

Note that despite us partly drawing the motivation for this work from a cooperative multi-vehicle scenario, the algorithm is evaluated in a distributed single-vehicle set-up. 
That is, a single vehicle localizes against a potentially remote cloud-based map-server, attempting to minimize the bandwidth usage while maintaining the localization performance. 
The method shown readily generalizes to and taps its full potential in a multi-vehicle set-up.

The remainder of this paper is structured as follows: 
In section \ref{sec:related_work}, our proposed selection method is put into context with other related work. 
Section \ref{sec:problem} and \ref{sec:landmark_ranking} derive the underlying appearance-based landmark ranking function the selection method is based upon, before an evaluation thereof is presented in section \ref{sec:evaluation}. 
To conclude, we summarize our findings and discuss future work in section~\ref{sec:conclusion}.

\section{Related Work}
\label{sec:related_work}
Extensive efforts have been made in the past years to adapt visual localization systems for long-term operation and resource-constrained environments. 
The methods presented in \cite{dayoub2008adaptive}, \cite{konolige2009towards}, \cite{milford2010persistent} all involve an adaptive selection of either landmarks or visual views in order to bound the growth of maps, while accounting for an environment subject to appearance change.
This selection may be based on a short-term/long-term memory model \cite{dayoub2008adaptive}, on clustering techniques \cite{konolige2009towards}, or on random pruning in neighbourhoods of high data density \cite{milford2010persistent}.
Similarly, the summary-mapping techniques proposed in \cite{muhlfellner2015summary} and \cite{dymczyk2015gist} aim at maintaining as compact and small a map representation as possible, while at the same time covering a high degree of variance in appearance.
All of these methods have in common, that the selection is an offline process performed prior to and/or independent of the robot's next operation. 
In contrast to that, our proposed selection method is an online process, selecting landmarks at runtime according to the prevailing appearance conditions, without modifying the underlying map.


In \cite{Linegar2015}, an online selection algorithm is presented that is, as in our case, adaptive to appearance conditions.
Rather than reasoning over relevant landmarks, different visual ``experiences" are prioritized for localization on resource constrained platforms. 
In contrast to this setting  based on ``experiences", all landmarks that we select are expressed in a common coordinate frame, which allows the poses of the vehicle to also be estimated in a common frame, independent of what landmarks are selected and hence what appearance condition the vehicle is exposed to. 
This enables a seamless integration of our method with other modules of an autonomous vehicle, such as planning, navigation, and control. 
Furthermore, by performing selection at the level of individual landmarks, our approach is more closely linked to the underlying environmental features. 
In this way, accounting for the fact that many landmarks may be shared among similar appearance conditions while others may be very distinct to certain conditions is implicitly handled by our framework. 

Landmark selection has also been studied in connection with specific tasks like path-planning and obstacle avoidance. The method presented by \cite{mu2015two} selects those landmark measurements from a map, which maximize the utility wrt. a predefined task, such as collision-free navigation. 
In contrast, the method we present selects landmarks based on the appearance condition the robot is exposed to during operation.

Recently, landmark co-occurrence statistics have been increasingly exploited in the context of place-recognition. In  \cite{cummins2011appearance} co-occurrence information is used to infer which types of features are often seen together, as this helps distinguishing places. 
Furthermore, in \cite{johns2013feature}, \cite{johns2014generative}, and \cite{stumm2015location} places are described and identified by constellations of visible landmarks or features grouped based on co-observability, therefore incorporating pseudo-geometric information in their representation.
Similarly, the works of \cite{li2010location} and \cite{mohan2015environment} rely on landmark co-occurence statistics for prioritizing relevant landmarks or environments for improved place-recognition efficiency. 
The clear correlation between the appearance of the environment and the co-observability of landmarks demonstrated in these works has inspired the selection algorithm presented in this paper. 
However, we propose to use the co-observability statistics in order to achieve a different goal, namely to infer which landmarks are likely to be observable in the near future during online operation, allowing to minimize data exchange.

Along that line, the work presented in \cite{carlevaris2012learning} is similar to ours, as they learn co-observability relationships across different appearance conditions in order to predict the current operating condition of the robot.
The main difference is that their co-observability prediction is performed on the level of camera images, whereas we propose to exploit co-observability on the level of individual 3D landmarks, contained in a sparse geometric visual map.

\section{Problem statement}
\label{sec:problem}
%
%
We consider a scenario in which iterative visual localization systems, such as the ones described in \cite{muhlfellner2015summary}, \cite{lategahn2012city} or \cite{churchill2013experience}, are used for periodic correction of pose estimates obtained from odometry.
The underlying map is assumed to be stored as a pose-graph in a multi-session SLAM framework, as described in \cite{cieslewski2015map}, which contains information about landmarks (position estimates and feature descriptors of respective observations).
Additionally, bundle adjustment and loop-closure \cite{lynen2014placeless} have been performed to merge identical landmarks observed in multiple mapping sessions, register the maps to each other and refine the resulting joint map.

%
%
Each of the mapping sessions that is used for generating the multi-session map may have been recorded at different times, with possibly very different appearance conditions.
Therefore, the observability of individual landmarks is highly variable and not all are equally useful for localization under a specific appearance condition.
%
In a scenario as described in section \ref{sec:introduction}, where the map is located on a cloud-based server, a decision has to be made about which landmarks to use, and hence transmit to the vehicle, for each localization attempt during online operation.
In order to support this decision, the vehicle provides the server with a rough initial pose estimate and information on which landmarks have recently been observed along the trajectory.
Based upon this information, we propose a landmark selection method aimed at only selecting those landmarks for transmission to the vehicle, which are deemed likely observable, and thus useful for localization.

In particular, we are interested in the following landmark ranking function:
\begin{equation}
\label{def_ranking_f}
f_{\poseguesslong{k}, \pastobslong}(\candidate) \coloneqq \Pb(\candidate \mid \poseguesslong{k}, \pastobslong)	
\end{equation}
It denotes the probability of observing landmark $\candidate$ at time $t_k$, given an initial estimate of the vehicle's pose denoted by $\poseguesslong{k}$, and a list of recently observed landmarks before time $t_k$ denoted by $\pastobslong$.
In order to improve readability, we use abbreviated symbols $\poseguess{k}$ and $\pastobs$ for the remainder of this section.

%
%

%
%
%
%
%
%
%
%

\section{Probabilistic landmark ranking}
\label{sec:landmark_ranking}
%
%
%
Using Bayes' rule, expression \ref{def_ranking_f} can be reformulated as 
\begin{equation}
	\Pb(\candidate \mid \poseguess{k}, \pastobs)	
	=\fr{\Pb(\pastobs \mid \candidate, \poseguess{k}) \cdot \Pb(\candidate \mid \poseguess{k})}{\Pb(\pastobs \mid \poseguess{k})}
\end{equation}
Probability $\Pb(\pastobs \mid \poseguess{k})$ is a fixed constant and does not influence the ranking of landmarks, whereas $\Pb(\candidate \mid \poseguess{k})$ denotes the pose dependent probability of observing landmark $\candidate$ at time $t_k$. 
We model the latter with a uniform distribution over all landmarks observed from within a given radius $r$ around the estimate of the vehicle's pose $\poseguess{k}$, and zero for other landmarks.
In practice, this allows retrieving an appearance-independent tight spatial subset of possibly observable candidate landmarks, denoted by $\mathcal{C}_k$, as described in \cite{muhlfellner2015designing}, which are then ranked according to $\Pb(\pastobs \mid \candidate, \poseguess{k})$.
\newline
%

%

%
%
%
%
%

%
%
The term $\Pb(\pastobs \mid \candidate, \poseguess{k})$ can be interpreted as the probability of having recently observed the set of landmarks $\pastobs$, given an estimate of the vehicle's current pose $\poseguess{k}$  and that landmark $\candidate$ is observed at time $t_k$.
Since past landmark observations are independent of the current vehicle's pose, we can reformulate as follows:
%
\begin{equation}
\label{def_rank_likelihood}
	\Pb(\pastobs \mid \candidate, \poseguess{k}) = \Pb(\pastobs \mid \candidate)
\end{equation}

%
%
%
%
%
%
%
%

\begin{figure}
\center
\includegraphics[height=120px]{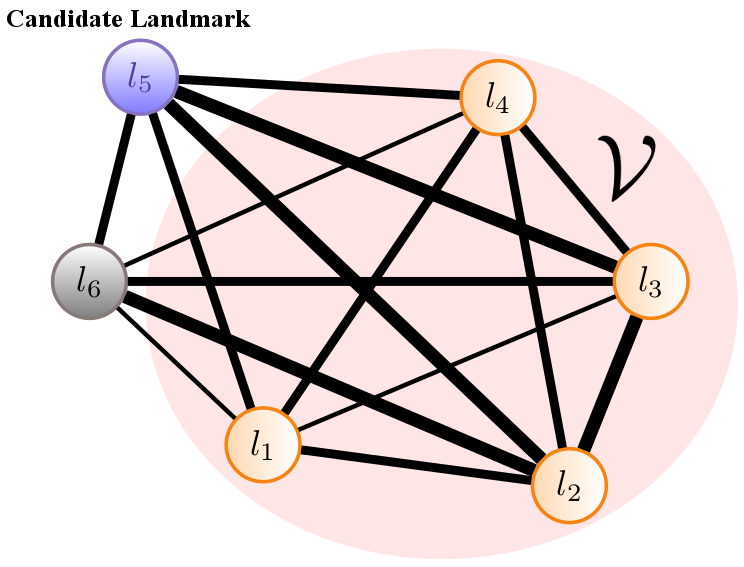}
\caption{The co-observability graph represents which landmarks (vertices) have been co-observed how often in the past (edges). From knowing which landmarks have recently been observed along the current traversal ($\mathcal{V}$, orange), our goal is to decide how likely a candidate landmark (blue) is to be observed at the current time-step.\label{fig:graph}}
\vspace{-3mm}
\end{figure}

%
%
From a frequentist's perspective, $\Pb(\pastobs \mid \candidate)$ could be approximated by the number of times, \textit{all} landmarks in $\pastobs$ and $\candidate$ have been observed together in the past, divided by how often $\candidate$ was observed.
For such a quantification to hold as a good approximation, the amount of co-observation data must be very high and no "appearance-outliers" (i.e. landmarks recently observed although they do not conform with the overall prevailing appearance condition) may be present in $\pastobs$ - two requirements unlikely met in practical applications.
We therefore propose to approximate $\Pb(\pastobs \mid \candidate)$ by explicitly accounting for limited statistical data and possible "appearance-outliers".

%
%
%

%
%
We assume the multi-session map has been generated from datasets representing traversals through the mapped area under different appearance conditions, possibly augmented with additional co-observation statistics from further traversals through the map.
Thus, we interpret the set of all past traversals, denoted by $\lochist$, as an enumeration over appearance conditions represented in the multi-session map.
%
%
%
With this, we can use the law of total probability in order to obtain the following decomposition:
\begin{equation}
\label{law_of_total_probs}
	\Pb(\pastobs \mid \candidate) = \sum_{z \in \lochist} \Pb(\pastobs \mid z, \candidate) \cdot \Pb(z \mid \candidate)
\end{equation}
We model $\Pb(z \mid \candidate)$ with a uniform distribution over all traversals $z$ in which $\candidate$ was observed, and zero for all other traversals.
For a traversal observing $\candidate$, the likelihood $\Pb(\pastobs \mid z, \candidate)$ becomes independent of $\candidate$, and we can thus reformulate \eqref{law_of_total_probs} as:
\begin{equation}
\label{law_of_total_probs_ref}
	\Pb(\pastobs \mid \candidate) = \frac{1}{\vert \lochist' \vert}\sum_{z \in \lochist'} \Pb(\pastobs \mid z)
\end{equation}
where $\lochist'$ denotes all traversals in $\lochist$ where $\candidate$ was observed in.
Due to the fact that this appearance term $\Pb(\pastobs \mid \candidate)$ is only evaluated for a spatially local subset of landmarks $\mathcal{C}$ (retrieved evaluating $\Pb(\candidate \mid \poseguess{k})$) and the appearance condition is assumed to be locally stable, both in a spatial and temporal manner, it suffices to consider a landmark as observed in traversal $z$ if it has been observed at least once along the traversal, regardless of the place or time. 
Analogously, two landmarks are considered co-observed in a traversal $z$, if both of them have been observed at least once in $z$, at potentially different times and places.

For each of these past traversals in $\lochist'$ either none, some or even all landmarks in $\pastobs$ were observed. 
To account for potential "appearance-outliers", we model the probability $\Pb(\pastobs \mid z)$, namely the probability of observing $\pastobs$ in traversal $z$, to be equal to the fraction of landmarks in $\pastobs$ actually observed in traversal $z$.
%
%
%
%
%
%

In conclusion, we can express our ranking function as follows:
\begin{equation}
\label{ranking_done}
f_{\poseguess{k}, \pastobs}(\candidate) = \frac{1}{\vert \lochist' \vert} \sum_{z \in \lochist'}  \vert \pastobs_{z} \vert
\end{equation}
where $\vert \pastobs_{z} \vert$ denotes the number of landmarks of $\pastobs$ that were observed in traversal $z$.
For simplicity, the constant denominator $\vert \pastobs \vert$ is omitted from the sum as it does not influence the ranking.

An intuitive graphical interpretation of this ranking function is shown in figure \ref{fig:graph}, where landmarks are represented as vertices and the co-observation relation as weighted edges connecting them.
The landmarks colored in orange denote the recent observations $\pastobs$, while the candidate landmark is colored in blue.
The score of candidate $\candidate$ according to the presented ranking function corresponds to the sum of co-observation connections into $\pastobs$, normalized by the total number of traversals observing $\candidate$. 
It represents how tightly a candidate is connected to the set $\pastobs$ in the pair-wise co-observation graph.
Hence, candidates with a strong connection into $\pastobs$ are favored over those with only a weak connection, relating to how likely the given candidates are co-observed with $\pastobs$.

\section{Evaluation}
\label{sec:evaluation}
The proposed landmark selection method exploits varying appearance conditions expressed in a single multi-session map of sparse landmarks.
In order to be able to build such multi-session maps,
sufficient data must be collected during the mapping phase, that is diverse enough to cover several different conditions, while exhibiting also some overlap in appearance. 
To the best of our knowledge, no publicly available datasets fulfill these criteria. 
We therefore evaluate our selection method in two complementary experimental scenarios using our own datasets recorded for the purpose of evaluating long-term visual localization and mapping.

In scenario A, a multi-session map of an open-space parking lot area is created, with datasets spanning over one year, covering the entire range of weather conditions and seasonal change. 
In scenario B, a city environment is mapped over the course of six hours from day-time to night, covering the most extreme change in appearance from daylight to night-time under artificial street lighting. 

\begin{figure}
\begin{subfigure}{.31\textwidth/2}
  \includegraphics[width=\linewidth]{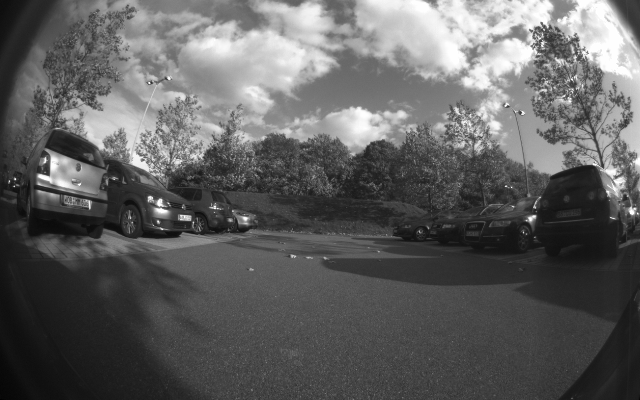}
  \caption{10/16/2013}
  \label{fig:sub1}
\end{subfigure}%
\begin{subfigure}{.31\textwidth/2}
  \includegraphics[width=\linewidth]{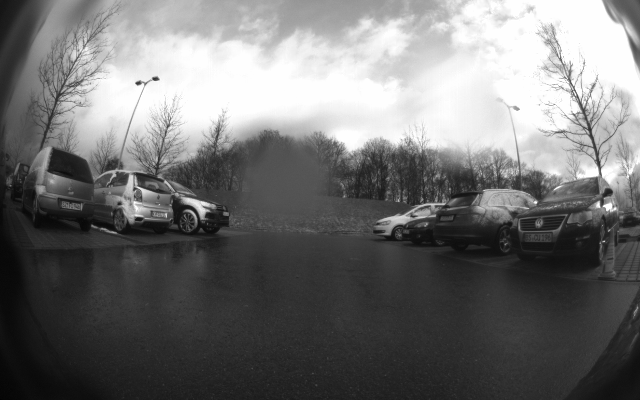}
  \caption{12/06/2013}
  \label{fig:sub2}
\end{subfigure}%
\begin{subfigure}{.31\textwidth/2}
  \includegraphics[width=\linewidth]{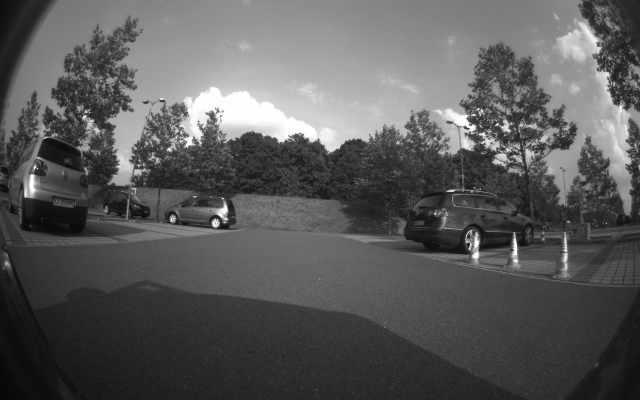}
  \caption{07/16/2014}
  \label{fig:sub2}
\end{subfigure}
\caption{Example images from the parking-lot environment, showing the varying appearance conditions induced by changes in lighting, weather, as well as foliage.}
\label{fig:example_parking}
\vspace{-2mm}
\end{figure}

\begin{figure}
\begin{subfigure}{.31\textwidth/2}
  \includegraphics[width=\linewidth]{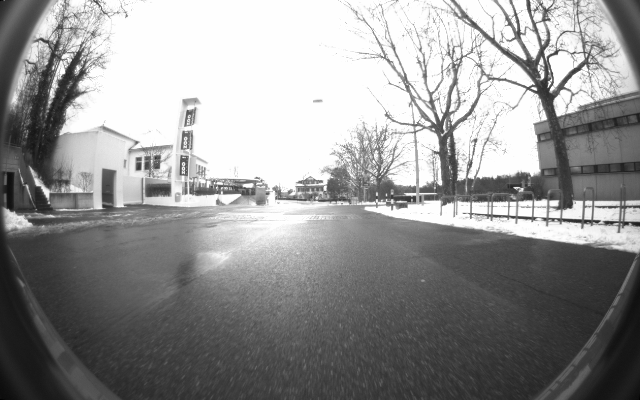}
  \caption{15:15}
  \label{fig:sub1}
\end{subfigure}%
\begin{subfigure}{.31\textwidth/2}
  \includegraphics[width=\linewidth]{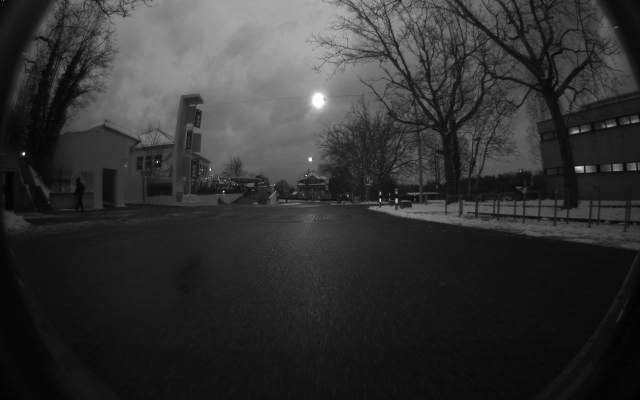}
  \caption{17:08}
  \label{fig:sub2}
\end{subfigure}%
\begin{subfigure}{.31\textwidth/2}
  \includegraphics[width=\linewidth]{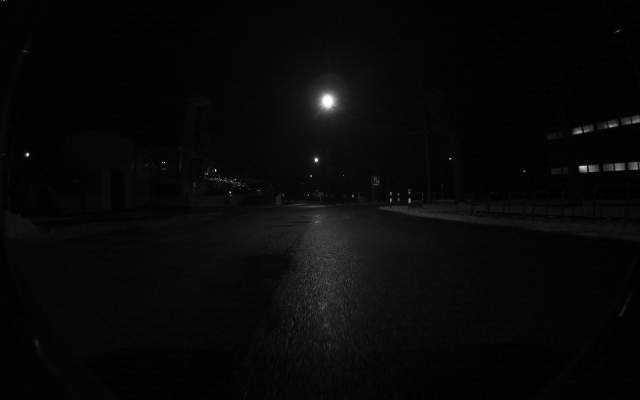}
  \caption{18:05}
  \label{fig:sub2}
\end{subfigure}
\caption{Example images from the city environment, showing the changes in appearance from day to night.}
\label{fig:example_city}
\vspace{-3mm}
\end{figure}

A total of $31$ traversals of the parking lot environment (roughly $155m$ each) and $26$ traversals of the city environment (roughly $455m$ each) were recorded, resulting in an accumulated driving distance of about $16.5 km$. 
For each environment, half of the recordings distributed over the respective time spans were used to build the map and augment the co-observability data, while the other half ($\approx 8km$) were used for the evaluation.
Example images from each of the two environments can be seen in figures~\ref{fig:example_parking} and~\ref{fig:example_city}.

The vehicle's sensor setup consists of four wide-angle fish-eye cameras - one in each cardinal direction - and wheel odometry sensors. The cameras run at a frame-rate of 12.5Hz. All images were recorded in gray-scale and down-scaled to 640px x 480px.
%

During each traversal, localization is performed iteratively.
For each image, a rough initial pose estimate is calculated (based on the previous pose estimate and integrated wheel odometry), a candidate set $\mathcal{C}_k$ is retrieved, from where a top-ranked subset of landmarks is selected yielding $\mathcal{S}_k$, landmark-keypoint matches are formed, the initial pose estimate is refined using a non-linear least-squares estimator, and a final match classification step distinguishes between inliers and outliers.
The landmarks associated with these inlier matches are considered the observed landmarks at a given time $t_k$, as described in section \ref{sec:problem}, and are denoted by $\mathcal{O}_k$.


In $\pastobs$, we only keep observed landmarks from the previous localization (i.e. from time $t_{k-1}$), since in our experimental scenarios, no significant improvement was observable when extending $\pastobs$ over a longer time window.





\subsection{Ranking function and selection policies}
We aim at demonstrating that with our selection method, we can significantly reduce the number of landmarks used for localization while simultaneously maintaining a similar localization performance. 
For this, we evaluate several performance metrics for three different selection policies: i) using the ranking function derived in section \ref{sec:problem} and \ref{sec:landmark_ranking}, ii) random selection, and iii) simply selecting all landmarks.
The latter marks the baseline for our experiments, while random selection constitutes a lower bound for the quality of our ranking-based selection. 



We formally define the selection policy as follows:
\[
\Omega(\mathcal{C}, f(), r, m) \coloneqq \text{Select } n \text{ top-ranked landmarks}
\]
where $n = min(r * \vert \mathcal{C} \vert, m)$, based on a selection ratio $r$ and a maximum number of landmarks $m$.
Consequently, $\Omega(\mathcal{C}, f(), 1.0, \infty)$ corresponds to selecting all landmarks.
While parameter $m$ directly relates to some fixed constraint on the available network bandwidth, the selection ratio $r$ prevents the algorithm to select poorly ranked landmarks in spatial locations of generally few visual cues (small $\vert \mathcal{C} \vert$).
For the sake of notational brevity, we abbreviate $\Omega(\mathcal{C}, f(), r, m)$ by $\Omega(f(), r, m)$ in the plots shown.
With $f_{rank}()$ we refer to the ranking function derived in section \ref{sec:problem} and \ref{sec:landmark_ranking}, while $f_{rand}()$ denotes a random uniform ranking across $\mathcal{C}$. 

\subsection{Metrics}
The following metrics are evaluated and respective experimental results are presented in the remaining subsections. 

\noindent \emph{a) Ratio between the number of \textit{selected} landmarks and the number of candidate landmarks}:
\[
r_{k}^{sel} \coloneqq \frac{\mid \mathcal{S}_k \mid }{\mid \mathcal{C}_k \mid}
\]

This metric directly relates to the amount of data transmission saved by performing landmark selection.

\noindent \emph{b) Ratio between the number of \textit{observed} landmarks with and without a selection at a given time $t_k$}: 
\[
r_{k}^{obs} \coloneqq \frac{\vert \mathcal{O}_k^{\Omega(f_{rank}(),r,m)} \vert }{\vert \mathcal{O}_k^{\Omega(f(), 1.0, \infty)} \vert }
\]
The number of observed landmarks constitutes a good indicator of the resulting pose estimate's accuracy (see \cite{muhlfellner2015summary}) and the ratio $r_{k}^{obs}$ is directly related to how well the selection predicts the current appearance condition. 
An ideal landmark selection method would achieve a ratio close to $1.0$, with a significantly reduced number of selected landmarks.

\noindent \emph{c) RMS errors for translation and orientation wrt. wheel-odometry}:
\newline
For each localization attempt, the transformation between the initial rough pose estimate, based on the visual pose estimate from $t_{k-1}$ and forward integrated wheel-odometry, and the refined visual pose estimate from $t_k$, can be computed, and is denoted by $T_{B_{k}^{est}B_{k}^{odo}}$.
Conceptually, this transformation corresponds to the odometry drift correction. 
While wheel-odometry accumulates drift over time, it is locally very smooth.
Since this refined visual pose estimate is only based on the positions of the matched landmarks, and in particular no odometry fusion is performed, the magnitude of $T_{B_{k}^{est}B_{k}^{odo}}$ is dominated by the uncertainty of the visual estimate. 

We compute separate RMS errors for both the translational and rotational component of $T_{B_{k}^{est}B_{k}^{odo}}$. 
%
%
Note that this metric does not describe the absolute localization accuracy. 
It only constitutes an indicator for the relative uncertainty of the visual pose estimates allowing a comparison between the three cases of selecting all landmarks, random selection, and ranking-based selection.
%
%
\begin{figure}
\includegraphics[width=\textwidth/2]{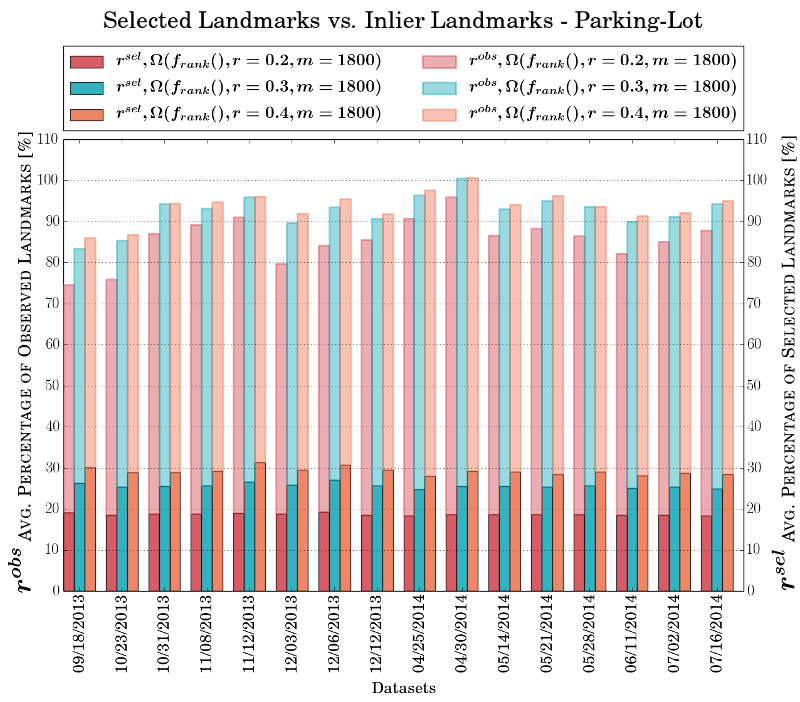}
\caption{Illustration of the relation between the average number of selected landmarks and the average number of observed landmarks for datasets from the parking-lot scenario. The lower bars (between $20$-$30\%$) correspond to the average percentage of selected landmarks $r^{sel}$, while the upper bars (between $75$-$100\%$) show the average percentage of observed landmarks $r^{obs}$.
\label{plot_bars_fofa}}
\vspace{-3mm}
\end{figure}

\begin{figure}
\includegraphics[width=\textwidth/2]{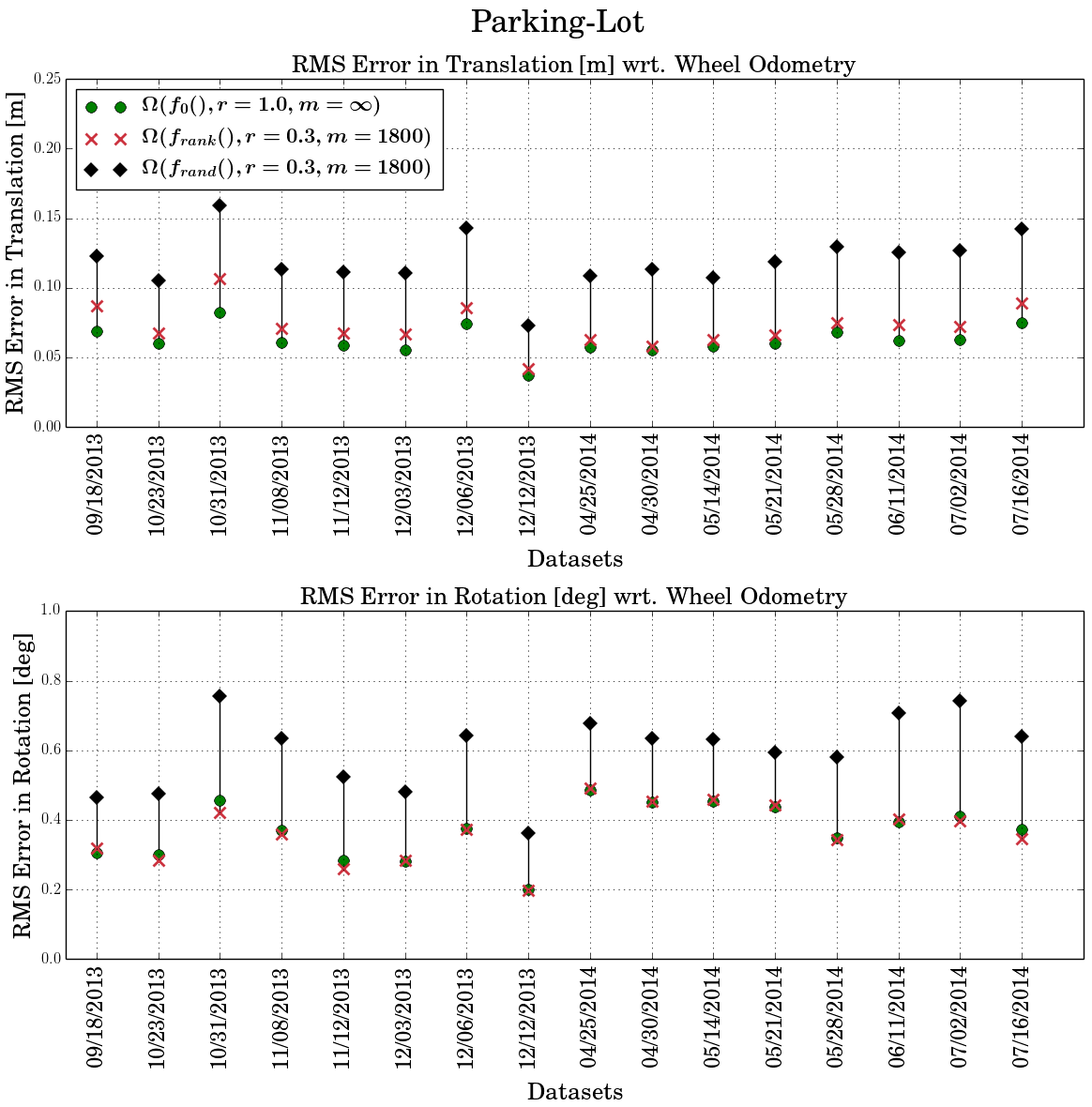}
\caption{RMS error in translation and rotation wrt. wheel odometry for the parking-lot scenario. In green, the RMS error is shown for the case where all landmarks are used, while black diamonds indicate results from random selection, and red crosses for the proposed ranking-based selection method.
\label{plot_accuracy_fofa}}
\vspace{-3mm}
\end{figure}
%
%
\subsection{Parking-Lot experiments}
Figure \ref{plot_bars_fofa} shows the relation between selected landmarks and observed landmarks for the parking-lot experiment. 
For each dataset, three different sets of selection policy parameters are evaluated, corresponding to more and less strict landmark selection.
While on average only $20$-$30\%$ of the total landmarks are used for localization, the ratio of observed landmarks with and without selection still remains between about $75$-$100\%$. 
For the dataset recorded on April 30\textsuperscript{th} 2014, the average $r^{obs}$ value even lies slightly above $100\%$. 
This is due to the fact that by eliminating landmarks inconsistent with the current appearance prior to the 2D-3D matching, the chance of wrong keypoint-landmark associations is reduced, potentially yielding even more observed landmarks in the case of ranking-based selection as compared to if all landmarks are selected.

\begin{figure}
\includegraphics[width=\textwidth/2]{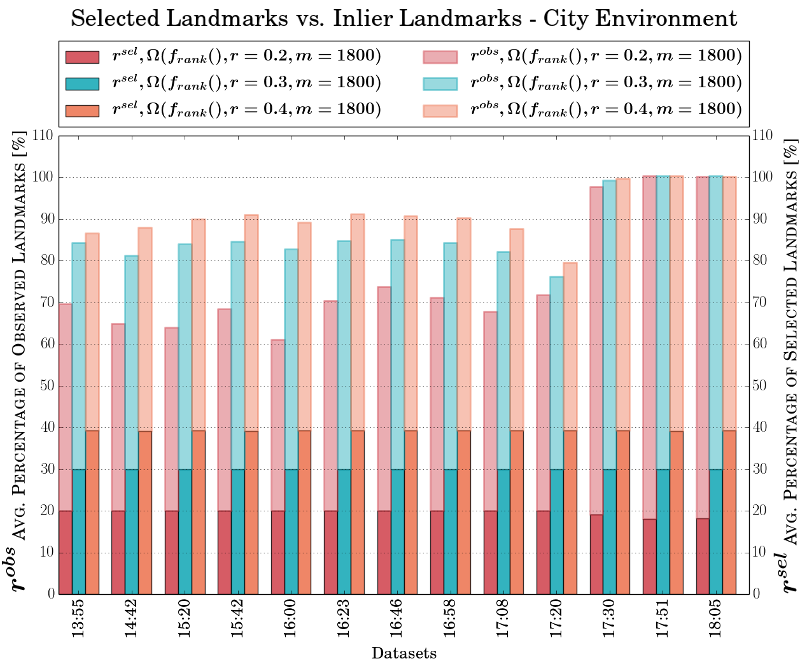}
\caption{Illustration of the relation between the average ratio of selected $r^{sel}$ and the average ratio of observed landmarks $r^{obs}$ for the city environment datasets. 
\label{plot_bars_zoo}}
\vspace{-4mm}
\end{figure}

In addition, figure \ref{plot_accuracy_fofa} shows the RMS error for translation and orientation for the three cases of using all landmarks, ranking-based selection, and random selection - the latter two with $r = 0.3$ and $m = 1800$.
From this plot, we see that the rotational component is mainly unaffected by the landmark selection, while there is a slight increase in the RMS error for the translational part.
In effect, the decrease in the number of observed landmarks results in a slightly less well constrained position estimate, whereas the orientation remains well constrained even with fewer observed landmarks. 
This is due to the fact, that, for a pure visual pose estimate, the translational component strongly depends on the spatial distribution of observed landmarks, especially on their distance from the vehicle, while the orientation does not.
However, the translational RMS error remains significantly lower than for the case of random selection, indicating meaningful landmark selection with the proposed ranking function.


\begin{figure}
\includegraphics[width=\textwidth/2]{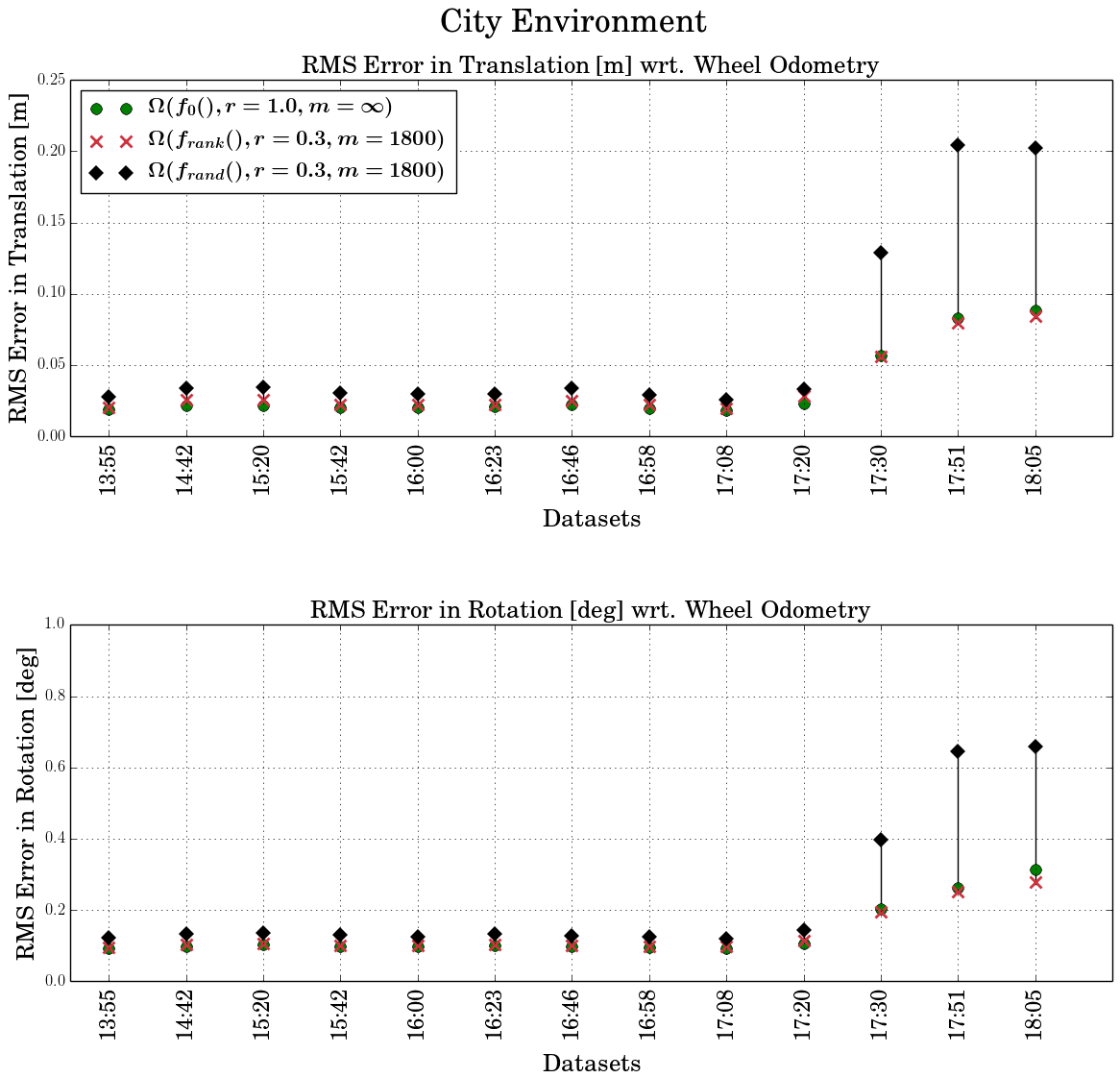}
\caption{RMS error in translation and rotation wrt. wheel odometry for the city environment scenario. Green corresponds to using all landmarks, while the black diamonds indicate results from random selection, and red crosses from the proposed ranking-based selection method.
\label{plot_accuracy_zoo}}
\vspace{-3mm}
\end{figure}

\subsection{City Environment experiments}
Figure \ref{plot_bars_zoo} again shows the relationship between the ratio of selected and observed landmarks, this time for the city environment. 
During daytime, an average observation ratio between $60\%$ and $90\%$ is achieved, depending on the strictness of selection, while at night-time, $100\%$ is reached almost independent of how many landmarks were selected. 
In contrast to the year-long parking-lot scenario with a high number of varying appearance conditions, in this scenario, we essentially have two very distinct conditions, namely day-time, and night-time, with a far greater total number of landmarks at day-time than at night-time. 
Therefore, selecting even as much as $40\%$ of the candidate landmarks at day-time may still exclude valid day-time landmarks, simply because of the limited number of selected landmarks.
At night-time, the opposite is true, where even a very strict selection of below $20\%$ allows selecting all relevant landmarks under this condition.
This effect is also well visible in the RMS error plots in figure \ref{plot_accuracy_zoo}.
At day-time, even a random selection performs relatively well, since the day-time landmarks are in vast majority. 
At night-time, however, our ranking function not only outperforms a random selection, but even achieves slightly better results than when all landmarks are selected. 
As already mentioned above, this is due to the reduced chance of forming wrong keypoint-landmark associations, allowing to achieve a more robust pose estimate.


\subsection{Shared vs. appearance-specific landmarks}
In order to demonstrate that our selection method favors different landmarks under different appearance conditions, we evaluate the pair-wise fraction of jointly selected landmarks between two datasets.

\begin{figure}
\includegraphics[width=\textwidth/2]{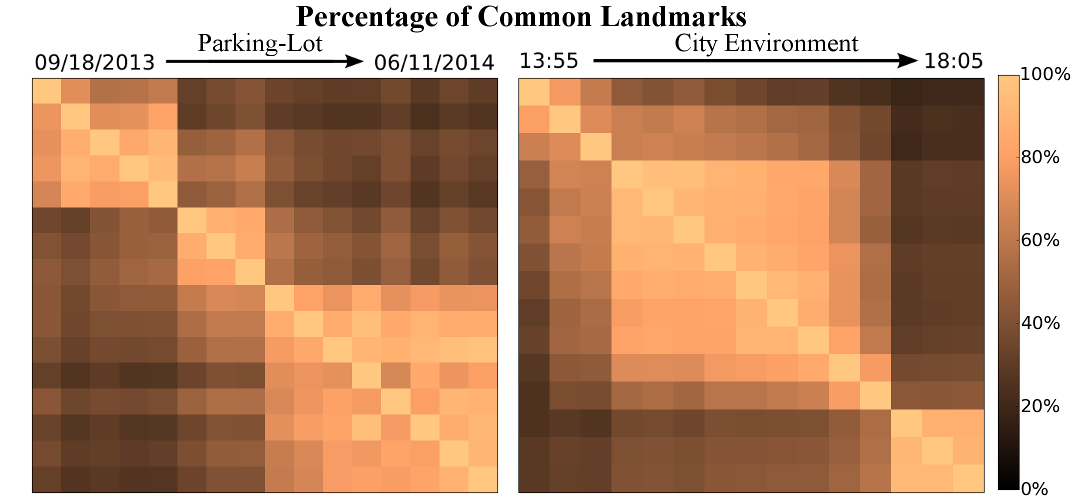}
\caption{Fraction of jointly selected landmarks between individual datasets.
\label{plot_foo_heatmap}}
\vspace{-6mm}
\end{figure}

The results are depicted in figure \ref{plot_foo_heatmap} for the two scenarios and a selection ratio $r = 0.25$ and maximum number of landmarks $m=1800$.

For the parking-lot scenario, a clear seasonal pattern can be observed, whereas for the city environment, a shift from day-time to night--time landmarks is visible.
About $10\%$ of the landmarks selected in any dataset are jointly selected in all datasets for the parking-lot scenario, whereas this fraction is as low as $2.5\%$ for the city-environment.


\section{Conclusion} 
\label{sec:conclusion}
We have presented an appearance-based landmark selection method for visual localization systems allowing to significantly reduce the data exchange during online operation between a vehicle and a cloud-based map server.
Using a simple ranking function, we can distinguish between landmarks that are useful and not useful for localization under the current appearance conditions, using co-observation statistics from previous traversals through the mapped area.
%
The selection method is evaluated in two environments undergoing long-term seasonal and weather change on the one side, and a full transition from day- to night-time on the other side, in combination covering a large extent of possible appearance variations for a visual localization system.
The number of landmarks used for localization under a specific appearance condition can be reduced to as little as $30\%$ while still achieving localization performance comparable to when all landmarks are used instead.
Importantly, in environments undergoing extreme changes in appearance with a clear association of landmarks to the appearance (e.g. day-time and night-time) a very precise selection is possible, even outperforming the case where all available landmarks are used.
However, the results of the day/night experiment further show that defining an appearance-independent number of landmarks to select at each time-step may not adequately account for the potentially very unbalanced number of landmarks useful under a certain appearance condition.
Therefore, in future work, more complex selection policies adapting the number of landmarks to select to the prevailing appearance condition ought to be investigated.
In addition to that, more sophisticated appearance outlier detection could further improve the results. 
Last but not least, extending our appearance-based ranking function with further aspects, such as the spatial distribution and uncertainty of landmark positions, and/or combining it with summary-map techniques such as the ones presented in \cite{muhlfellner2015summary} or \cite{dymczyk2015gist}, could significantly boost the performance and yield better localization accuracy with even fewer selected landmarks.

\section*{ACKNOWLEDGMENT}

This project has received funding from the European Union’s Horizon 2020 research and innovation programme under grant agreement No 688652 and from the Swiss State Secretariat for Education, Research and Innovation (SERI) under contract number 15.0284.


\bibliographystyle{IEEEtran}
\bibliography{bibliography/references_short}

\end{document}